%% file: neurips_2026.tex
\documentclass{article}

\usepackage[main, final]{neurips_2026}
\usepackage[utf8]{inputenc} % allow utf-8 input
\usepackage[T1]{fontenc}    % use 8-bit T1 fonts
\usepackage{url}            % simple URL typesetting
\usepackage{hyperref} 
\usepackage{booktabs}       % professional-quality tables
\usepackage{amsfonts}       % blackboard math symbols
\usepackage{nicefrac}       % compact symbols for 1/2, etc.
\usepackage{microtype}      % microtypography
\usepackage{xcolor}         % colors
\usepackage{textgreek} 
\usepackage{graphicx} 
\usepackage{caption}
\usepackage{amsmath}
\usepackage{multirow}
\usepackage{float}
\usepackage{algorithm}
\usepackage{algpseudocode}

\title{SAM 3D Animal: Promptable Animal 3D Reconstruction from Images in the Wild}

\author{%
    \textbf{Xuyi Hu}$^{1^{*}}$ \quad
    \textbf{Jin Lyu}$^{2^{*}}$ \quad
    \textbf{Jiuming Liu}$^{1}$ \quad
    \textbf{Yebin Liu}$^{3}$ \quad
    \textbf{Silvia Zuffi}$^{4}$ \quad
    \textbf{Liang An}$^{5\dagger}$ \quad
    \textbf{Stefan Goetz}$^{1}$
    \\[0.8em]
    \normalsize
    $^{1}$University of Cambridge \quad
    $^{2}$Southern University of Science and Technology \quad
    $^{3}$Tsinghua University \\
    \normalsize
    $^{4}$IMATI-CNR, Milan, Italy \quad
    $^{5}$Nanjing University
}
\begin{document}

\begingroup
\renewcommand{\thefootnote}{\fnsymbol{footnote}}
\footnotetext[1]{Equal contribution.}
\footnotetext[2]{Corresponding author.}
\endgroup

\maketitle

% \begin{figure*}[t] \centering
%     \includegraphics[width=1\textwidth]{images/teaser.pdf}
%     \caption{SAM3D Animal Model Structure. \label{fig:teaser}}
%     \vspace{-0.3cm}
% \end{figure*}

\begin{center}
    \includegraphics[width=\textwidth]{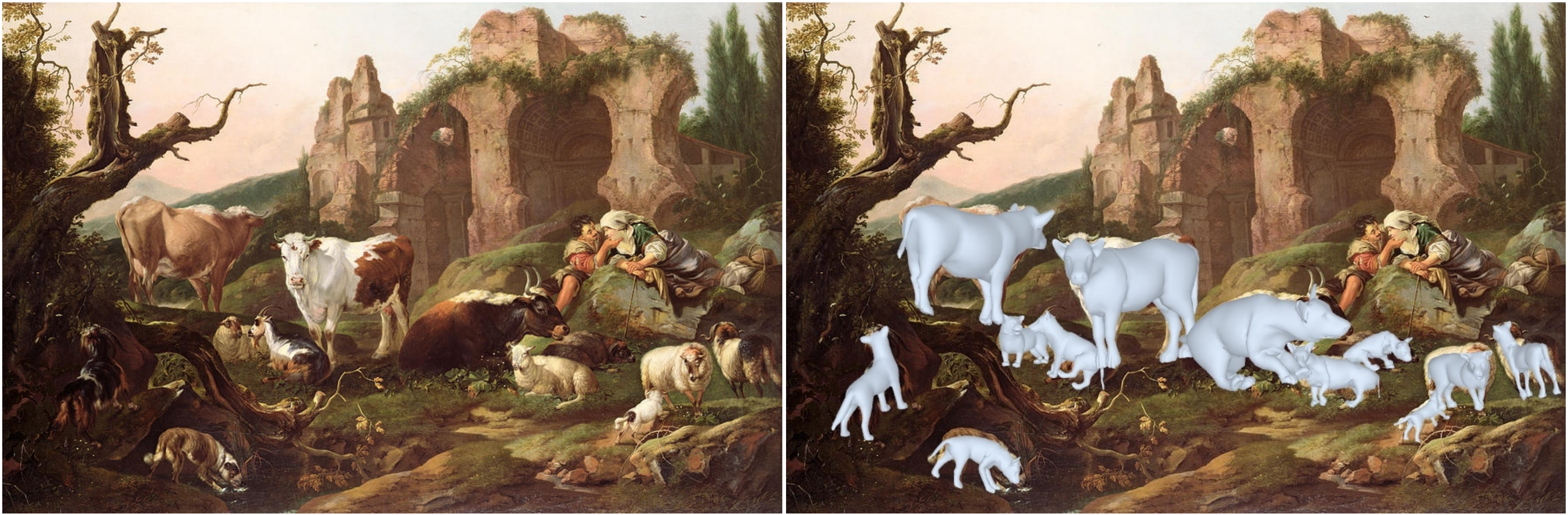}
    \captionof{figure}{A promptable view of multi-animal 3D reconstruction. We present \textbf{SAM 3D Animal}, a promptable framework that addresses the problem of joint 3D reconstruction of multiple animals from a single image. Here, we show the input image together with the overlay reconstruction results. \label{fig:teaser}}
\end{center}

% \begin{center}
%     \includegraphics[width=\textwidth]{images/teaser.pdf}
    
%     \small{Figure 1. SAM3D Animal Model Structure.}
% \end{center}
% --------------------------------------------------------------
% Include section files
% --------------------------------------------------------------
\input{sec/00_abstract}
\input{sec/01_introduction}
\input{sec/02_related_works}
\input{sec/03_method}

\input{sec/04_dataset}

\input{sec/05_experiment}

\input{sec/06_conclusion}

% --------------------------------------------------------------
% References
% --------------------------------------------------------------
\clearpage
\bibliographystyle{plain}
\bibliography{main.bib}

% --------------------------------------------------------------
% Supplementary Material
% --------------------------------------------------------------
\clearpage
\appendix
\input{sec/X_suppl}

\clearpage
\input{checklist}
\end{document}

%% file: sec/00_abstract.tex
\begin{abstract}
% \keywords{3D Reconstruction \and Animal Pose \and Single-Image 3D}

3D animal reconstruction in the wild remains challenging due to large species variation, frequent occlusions, and the prevalence of multi-animal scenes, while existing methods predominantly focus on single-animal settings. We present SAM 3D Animal, the first promptable framework for multi-animal 3D reconstruction from a single image. Built on the SMAL+ parametric animal model, our method jointly reconstructs multiple instances and supports flexible prompts in the form of keypoints and masks which enable more reliable disambiguation in crowded and occluded scenes. To train such a model, we further introduce Herd3D, a multi-animal 3D dataset containing over 5K images, designed to increase diversity in species, interactions, and occlusion patterns. Experiments on the Animal3D, APTv2, and Animal Kingdom datasets show that our framework achieves state-of-the-art results over both existing model-based and model-free methods, demonstrating a scalable and effective solution for prompt-driven animal 3D reconstruction in the wild. Project page: \url{https://georgehux.com/SAM3D-Animal-project-page/}

\end{abstract}

%% file: sec/01_introduction.tex
\section{Introduction}
% 2026.2.7 20:05 Jin Lyu
% Motivation: 
%   1. Lack of Single Image One Shot Multiple Animals Reconstruction (AniMer, Fauna ...).
%   2. Lack of 3D Multiple Animals Annotations Dataset (Animal3D, CtrlAni3D, and GenZoo are all single animal dataset, APT36K and Animal Kingdom are 2D dataset.)
% Novelty:
%   1. The first multiple 3D animal reconstrution method doesn't need bbox cropping.
%   2. The first 3D multiple animal dataset.
%   3. Supporting keypoints and mask prompts.

\label{sec:intro}
Animals are a fundamental part of the visual world, yet 3D reconstruction research remains heavily skewed toward humans. Human-centric methods have advanced pose and shape estimation dramatically~\cite{kanazawa2018end, zhang2021pymaf, zhang2023pymaf, goel2023humans, wang2024tram, baradel2024multi, li2026anylift}.
In contrast, animal reconstruction still suffers from scarce datasets, broad species variation, and inconsistent anatomical definitions.

Parametric animal models such as SMAL~\cite{zuffi20173d} and SMAL+~\cite{zuffi2024awol} provide an effective basis for recovering 3D pose and shape from a single image~\cite{zuffi2018lions,xu2023animal3d,niewiadomski2025generative,lyu2025animer,an2025animer+}. These approaches typically focus on one animal at a time and often rely on pre-cropped inputs or strong object detections. However, many in-the-wild animal scenes contain multiple individuals with mutual occlusion, and complex interactions that invalidate single-animal assumptions.

Multi-animal 3D reconstruction raises unique challenges beyond those of the single-object case. First, instance association becomes ambiguous when animals overlap or occlude one another. Second, pose and shape estimation must be jointly consistent across multiple hypotheses, since mistakes on one individual can be amplified by false depth ordering or incorrect occlusion reasoning. Third, available datasets rarely provide dense multi-animal 3D annotations, which hinders supervised learning for crowded scenes.

To overcome these challenges, we draw inspiration from promptable reconstruction in human vision. Recent works such as SAM 3D Body~\cite{yang2025sam3dbody} demonstrates that explicit prompts can guide a model to focus on a desired subject and resolve ambiguity in cluttered scenes. Prompts can take the form of keypoints or masks, each providing a different level of spatial and semantic guidance. 

In this paper, we introduce \textbf{SAM 3D Animal}, the first promptable framework for multi-animal 3D reconstruction, see Fig.~\ref{fig:teaser}. Our model uses the SMAL+ template~\cite{zuffi2024awol} and can ingest optional prompts in two modalities: keypoints for skeletal alignment and masks for precise silhouette discrimination. This promptable design allows SAM 3D Animal to recover multiple animals jointly from a single image. 
Different from SAM 3D Body, which reconstructs a single prompted subject per forward pass, our model adopts a set-prediction paradigm that recovers all animal instances in one shot via DETR-style~\cite{carion2020end} bipartite matching, eliminating the need for per-instance bounding-box cropping.

However, training such a multi-instance model with only 2D-annotated datasets is insufficient, as 2D keypoints and silhouettes alone cannot provide the per-instance 3D shape supervision needed to resolve inter-animal occlusions. To address this, we propose Herd3D, a multi-animal 3D dataset containing over 5K images with per-instance ground-truth meshes, designed to increase diversity in species, interactions, and occlusion patterns. The generation pipeline of Herd3D is adapted from GenZoo~\cite{niewiadomski2025generative}, where each animal is synthetically generated to produce an image with an image-aligned SMAL+ model.

To demonstrate the effectiveness of SAM 3D Animal, we compare it with state-of-the-art animal mesh recovery methods on publicly available datasets including Animal3D~\cite{xu2023animal3d}, APTv2~\cite{yang2023aptv2}, and Animal Kingdom~\cite{ng2022animal}. Even without any prompt, our model achieves competitive or superior results compared to existing methods. When prompts are provided, performance improves consistently across all benchmarks, with up to 54\% AP gain and 80\% mAP gain on the out-of-domain Animal Kingdom dataset over the strongest baseline, as well as 5.2 PA-MPJPE improvement on Animal3D. Ablation studies confirm that Herd3D brings consistent improvements, particularly on multi-animal benchmarks, and that keypoint prompts are the dominant contributor among prompt modalities, with performance scaling monotonically as the number of keypoints increases.

% Our main contributions are:
% \begin{itemize}
%     \item We propose SAM 3D Animal, the first promptable multi-animal 3D mesh reconstruction framework that recovers all animal instances from a single image via DETR-style set prediction, without per-instance bounding-box cropping.
%     \item We introduce Herd3D, a synthetic multi-animal 3D dataset with over 5K images across 118 species and per-instance ground-truth SMAL+ meshes, filling the gap of multi-animal 3D annotations for training and evaluation.
%     \item Extensive experiments show that our promptable design yields scalable improvements with prompt fidelity, achieving state-of-the-art performance on three benchmarks.
% \end{itemize}

%% file: sec/02_related_works.tex
\section{Related Work}
\subsection{Model-Free Reconstruction}
Model-free animal reconstruction learns 3D structure directly from image or video collections without assuming a predefined template~\cite{goel2020shape,wu2021rendering}. Early methods model category-specific articulated animals from single-view image collections by separating a predefined skeleton prior from instance-specific deformations~\cite{yao2022lassie,wu2023magicpony,wu2023dove,jakab2024farm3d}. Later approaches extend to a wider variety of species, either by learning a unified shape model~\cite{li2024learning} or by applying linear skinning to deform learned 3D object shapes~\cite{aygun2024saor}. However, these methods still struggle with extreme poses, heavy occlusions, and limited viewpoint coverage, often producing geometrically ambiguous reconstructions.

\subsection{Model-Based Reconstruction}
% Model-based reconstruction has achieved remarkable success in the human domain, where parametric models such as SMPL~\cite{loper2023smpl} provide a compact yet expressive representation of human body shape and pose. The success of these methods relies on two key ingredients: a well-designed articulated deformable model and large-scale manual annotations for training pose regressors~\cite{rajasegaran2022tracking,goel2023humans,shin2024wham,kocabas2021pare,kocabas2020vibe}. 

Model-based animal reconstruction typically relies on predefined quadruped templates such as SMAL~\cite{zuffi20173d} that encode the shape and articulation structure of specific animal families. These models either fit predefined 3D templates to animal images using 2D observations such as keypoints or silhouettes~\cite{zuffi2018lions,biggs2018creatures,borycki2026smal}, or directly reconstruct the 3D shape from image or video observations~\cite{cashman2012shape,yang2021viser,yao2022lassie,zuffi2024varen,lyu20264dequine,cho2025dogrecon}. This parametric formulation offers interpretable and controllable representations, which make the reconstructed animals readily animatable and editable.
Recent works further extend SMAL-based reconstruction to broader quadruped species and training settings. 
% BITE~\cite{ruegg2023bite} learns a dog-specific parametric prior, 
AWOL~\cite{zuffi2024awol} maps CLIP-style embeddings to the SMAL$^{+}$ parameter space for language- and image-guided animal shape generation. RAW~\cite{kulits2025reconstructing} reconstructs animals jointly with their surrounding environment, including multi-animal scenes. However, it relies on rigid animal assets rather than articulated animal models, and therefore does not address fine-grained articulated animal reconstruction.

% AniMer~\cite{lyu2025animer} improves pose and shape estimation with a family-aware Transformer, and GenZoo~\cite{niewiadomski2025generative} provides a scalable pipeline for generating realistic SMAL-based animal training data. 

% More recently, there have been works that extend the SMAL template to support a broader range of quadruped species. BITE~\cite{ruegg2023bite} expands 3D dog pose and shape estimation by learning a dog-specific parametric body prior (D-SMAL). AWOL~\cite{zuffi2024awol} maps CLIP-style language embeddings to the parameter space of SMAL$^{+}$ so that text and images can directly control and generate novel animal shapes without manual parameter tuning. RAW~\cite{kulits2025reconstructing} further enables joint reconstruction of animals and their surrounding environment. 
% AniMer~\cite{lyu2025animer} proposes a systematic framework for accurate animal pose and shape estimation by exploiting a family-aware Transformer. 
% GenZoo~\cite{niewiadomski2025generative} performs a scalable pipeline that samples diverse quadruped pose and SMAL-based shape and then generates realistic animal images for training 3D animal reconstruction models.  

\subsection{Animal Pose Estimation Datasets}
In comparison to humans, the construction of large-scale animal datasets is significantly more challenging because animals are difficult to capture in controlled environments and exhibit substantial morphological diversity across species. Animal benchmarks, such as Stanford Extra~\cite{biggs2020left}, Animal Pose~\cite{cao2019cross}, and AwA-Pose~\cite{banik2021novel}, remain limited to 2D annotations. Existing 3D animal datasets, such as Animal3D~\cite{xu2023animal3d} and GenZoo~\cite{niewiadomski2025generative} predominantly focus on single-animal instances. 
Additionally, multi-view quadrupedal animal datasets focus mostly on single species such as dogs~\cite{peng2026interpet4d,wang2025dogmo,kearney2020rgbd}, horses~\cite{li2024poses}, and some agriculture animals like pigs~\cite{an2023three}, goats~\cite{jin2026monocular} or cattle~\cite{noronha2026moocap}. Meanwhile, real-captured dynamic datasets targeting wildlife animals such as WildDepth~\cite{aamir2026wilddepth} and AiM~\cite{zhaoweb}, and sim-to-real datasets for diverse species~\cite{lyu20264dequine,liu2026mozoo,martini2026bigmaq,artemis,li2024dessie}, all focus on single instances.
Previous large-scale social animal benchmarks like APT-36K~\cite{yang2022apt} and Animal Kingdom~\cite{ng2022animal} provide only 2D annotations. This limitation restricts the development of methods that can jointly model inter-animal occlusions, spatial relationships, and pose dependencies in multi-animal scenes. 

\subsection{Promptable Mesh Reconstruction}
% PromptHMR - SAM3D Body - Fast SAM3D Body - SAM4D Body
Promptable mesh reconstruction has recently emerged in human mesh recovery, where auxiliary cues guide 3D estimation under occlusion and crowding. PromptHMR~\cite{wang2025prompthmr} incorporates full-image context with spatial and semantic prompts for pose and shape estimation. SAM 3D Body~\cite{yang2025sam3dbody} extends this idea to full-body recovery through a promptable encoder-decoder architecture supporting keypoint and mask prompts. SAM-Body4D~\cite{gao2025sam} further leverages temporally consistent masklets to produce coherent mesh trajectories from videos. However, these methods are designed for humans, whereas our work targets the animal domain, where large morphological diversity, inter-animal occlusion, and multi-instance interactions must be jointly considered.

%% file: sec/03_method.tex
\section{SAM 3D Animal Model}
\subsection{Preliminary} 
\noindent\textbf{SMAL+.} SMAL+~\cite{zuffi2024awol}, denoted as \(\mathcal{M}(\boldsymbol{\beta}, \boldsymbol{\theta}, \boldsymbol{\gamma})\), extends the original SMAL~\cite{zuffi20173d} model by incorporating training samples from D-SMAL~\cite{rueegg2022barc} and hSMAL~\cite{li2021hsmal}, alongside new species such as the giraffe, bear, mouse and rat. This results in a broader, 145-dimensional shape space learned from a total of 145 animals. The inputs to SMAL+ are the shape parameters \(\boldsymbol{\beta} \in \mathbb{R}^{145}\) and the pose parameters \(\boldsymbol{\theta} \in \mathbb{R}^{35 \times 3}\) (using an axis-angle representation) and the global translation \(\boldsymbol{\gamma} \in \mathbb{R}^{3}\). By applying linear blendshapes and Linear Blend Skinning (LBS), SMAL+ outputs a posed mesh with vertices \(\mathbf{V} \in \mathbb{R}^{3889 \times 3}\) and faces \(\mathbf{F} \in \mathbb{N}^{7774 \times 3}\). 

\subsection{End-to-end Multi-Instance Network}

Given an animal image, our model can reconstruct all animals in the image without requiring preprocessed bounding boxes, and support masks or keypoints as prompts.

\begin{figure*}[t] 
    \centering
    \includegraphics[width=1\textwidth]{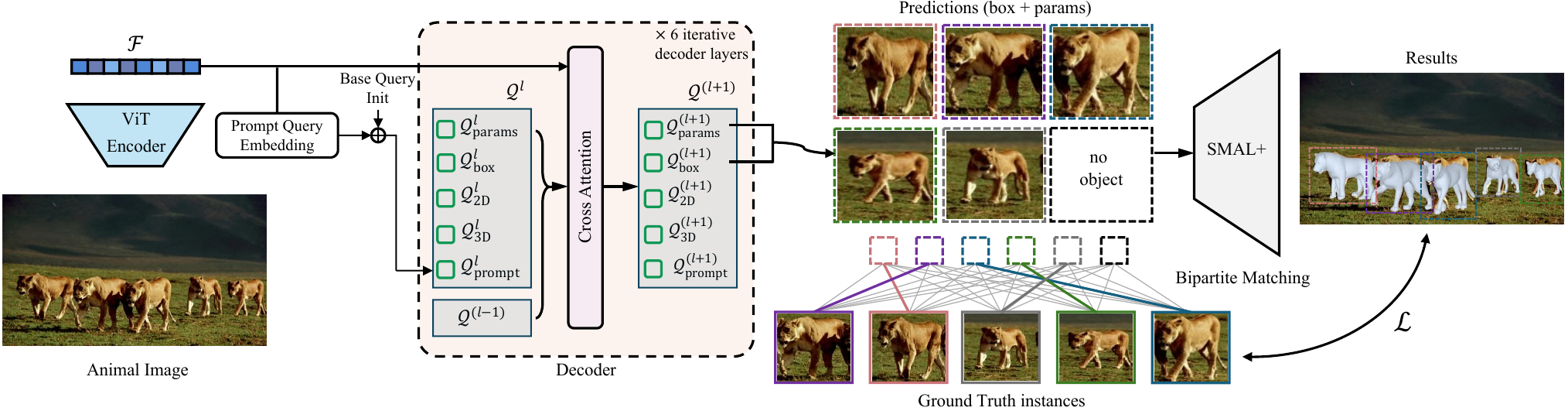}
    \caption{SAM 3D Animal model structure. 
    \label{fig:SAM3D_Animal}}
    \vspace{-0.3cm}
\end{figure*}

\noindent\textbf{Encoder.}
Starting from the image \(\boldsymbol{x} \in \mathbb{R}^{H \times W \times 3}\), we utilize the ViT-Huge Encoder~\cite{dosovitskiy2020image} to generate the feature tokens \(\mathcal{F} \in \mathbb{R}^{({H_{0} \times W_{0}}) \times C_{0}}\), where \(C_0, H_0\), and \(W_0\) are the channels, the height and width of the feature map, respectively. In our case, $W=H=512$, $C_0=1280$, $W_0=H_0=32$. 

\noindent\textbf{Decoder.} Inspired by SAM 3D Body~\cite{yang2025sam3dbody}, the decoder is a SAM-style promptable Transformer (see Fig.~\ref{fig:SAM3D_Animal}). Specifically, it takes the feature tokens \(\mathcal{F}\) and a set of query tokens as input, then performs cross-attention, and finally predicts the SMAL+ parameters, cameras, and bounding boxes. Note that, different from SAM 3D Body which predicts single person at a time, our model directly predict $P=30$ possible instances at a time, eliminating the need for bounding box input. 
The query tokens consist of five distinct token groups at state index $l=0,\ldots,L$, where $L=6$ and $l=0$ denotes the decoder input:
\begin{equation}
    \label{equ:decoder_tokens}
    \mathcal{Q}^{(l)} = \text{Concat}(\mathcal{Q}_{\text{params}}^{(l)}, \mathcal{Q}_{\text{box}}^{(l)}, \mathcal{Q}_{\text{2D}}^{(l)}, \mathcal{Q}_{\text{3D}}^{(l)}, \mathcal{Q}_{\text{prompt}}^{(l)}) \in \mathbb{R}^{N \times D},
\end{equation}
where \(\mathcal{Q}_{\text{params}}^{(l)}\), \(\mathcal{Q}_{\text{box}}^{(l)}\), \(\mathcal{Q}_{\text{2D}}^{(l)}\), and \(\mathcal{Q}_{\text{3D}}^{(l)}\) represent the SMAL+ pose tokens, bounding box tokens, 2D keypoint tokens, and 3D keypoint tokens, respectively, while \(\mathcal{Q}_{\text{prompt}}^{(l)}\) represents the interaction prompt tokens. 
The four base query groups, \(\mathcal{Q}_{\mathrm{params}}^{(0)}\), \(\mathcal{Q}_{\mathrm{box}}^{(0)}\), \(\mathcal{Q}_{\mathrm{2D}}^{(0)}\), and \(\mathcal{Q}_{\mathrm{3D}}^{(0)}\), are learnable embeddings. They are randomly initialized once before training and subsequently optimized jointly with the network. At inference time, their learned values are used without reinitialization. In contrast, \(\mathcal{Q}_{\mathrm{prompt}}^{(0)}\) is input-dependent and constructed from the keypoint prompts provided for the current image, rather than being a randomly initialized base query.
A keypoint prompt \(p_i=(u_i,v_i,c_i)\) comprises normalized coordinates \((u_i,v_i)\in[0,1]^2\) and a label: an Animal3D keypoint ID \(0\)--\(25\), a negative point \(-1\), or an invalid, missing, or padded point \(-2\). Its token is
\[
q_{\mathrm{prompt},i}^{(0)}=W_p e_i+b_p,\qquad
 e_i=\begin{cases}
\operatorname{PE}(u_i,v_i)+E(c_i), & c_i\in\{0,\ldots,25\},\\
E_{\mathrm{neg}}, & c_i=-1,\\
E_{\mathrm{invalid}}, & c_i=-2,
\end{cases}
\]
where \(\operatorname{PE}\) denotes positional encoding, \(E(c_i),E_{\mathrm{neg}},E_{\mathrm{invalid}}\) are learned label embeddings, and \(W_p,b_p\) are learned projection parameters. Prompts enter the first decoder layer through \(\mathcal{Q}^{(0)}=\operatorname{Concat}(\mathcal{Q}_{\mathrm{base}}^{(0)},\mathcal{Q}_{\mathrm{prompt}}^{(0)})\), with \(\mathcal{Q}_{\mathrm{base}}^{(0)}\) collecting the four base query groups. The feature dimension is \(D=1024\). At inference, \(N=55P+N_{\mathrm{prompt}}=1650+N_{\mathrm{prompt}}\): each hypothesis has one parameter, one previous-estimate, one box, and 26 tokens each for 2D and 3D keypoints; \(N_{\mathrm{prompt}}\) counts prompt tokens including padding.

Decoder block $l+1$ maps $\mathcal{Q}^{(l)}$ to $\mathcal{Q}^{(l+1)}$ (Fig.~\ref{fig:SAM3D_Animal}):
\begin{equation}
    \label{equ:cross_attn}
    \mathcal{Q}^{(l+1)} = \mathcal{D}_{l+1}(\mathcal{Q}^{(l)}, A^{(l)}, \mathcal{F}), \qquad l=0,\ldots,L-1,
\end{equation}
Here, $\mathcal{D}_{l+1}$ comprises self-attention, image cross-attention, and a feed-forward network; $A^{(l)}$ denotes query positional augmentation. Fig.~\ref{fig:SAM3D_Animal} highlights cross-attention. Its lower $\mathcal{Q}^{(l-1)}$ label schematically denotes previous-estimate conditioning, encoded once before decoding (using the initial estimate if none is supplied); it does not index a decoder layer.

For $l=0,\ldots,L-2$, SMAL+ and camera predictions yield projected 2D coordinates $\hat{x}_{\mathrm{2D}}^{(l+1)}$ and pelvis-centered 3D coordinates $\bar{x}_{\mathrm{3D}}^{(l+1)}$. Feedback updates the next input in place:
\begin{equation}
    \label{equ:2d_keypoints}
    \begin{aligned}
    \mathcal{Q}_{\mathrm{2D}}^{(l+1)} &\leftarrow \mathcal{Q}_{\mathrm{2D}}^{(l+1)} + \phi_{\mathrm{feat}}(\mathcal{F}(\hat{x}_{\mathrm{2D}}^{(l+1)})),\\
    A_{\mathrm{2D}}^{(l+1)} &\leftarrow m^{(l+1)}\odot\phi_{\mathrm{pos}}(\hat{x}_{\mathrm{2D}}^{(l+1)}),
    \end{aligned}
\end{equation}
\begin{equation}
    \label{equ:3d_keypoints}
    A_{\mathrm{3D}}^{(l+1)} \leftarrow \psi_{\mathrm{pos}}(\bar{x}_{\mathrm{3D}}^{(l+1)}).
\end{equation}
Here, $m^{(l+1)}$ selects in-frame projections with valid positive depth, $\mathcal{F}(\hat{x}_{\mathrm{2D}}^{(l+1)})$ denotes masked bilinear samples, and $\phi_{\mathrm{feat}},\phi_{\mathrm{pos}},\psi_{\mathrm{pos}}$ are learned mappings. Eqs.~(3)--(4) follow Eq.~(2), producing $(\mathcal{Q}^{(l+1)},A^{(l+1)})$ for the next block; 3D token contents and unlisted components remain unchanged. Fig.~\ref{fig:SAM3D_Animal} omits these updates. The sixth block skips feedback and generates final predictions from its parameter and box groups; ``params'' includes SMAL+ and camera parameters. Algorithm~\ref{alg:sam3d_forward} in the Appendix details the complete inference forward pass.

% \subsubsection{Bipartite Matching for Multi-Animal Instances}
\noindent\textbf{Bipartite Matching for Multi-Animal Instances.}
To enable end-to-end training without heuristic post-processing such as Non-Maximum Suppression (NMS), we adopt a set prediction formulation following the DETR paradigm~\cite{carion2020end}. Specifically, we employ bipartite matching via the Hungarian algorithm~\cite{kuhn1955hungarian} to find the optimal one-to-one assignment between the fixed-size set of $P$ predicted animal hypotheses and the $M$ ground-truth instances. The matching cost is a weighted combination of bounding box $\mathcal{L}_1$ distance, Generalized IoU~\cite{rezatofighi2019generalized}, focal-style confidence penalty~\cite{Su_2025_CVPR}, and masked 2D keypoint distance. Once the optimal assignment is established, predicted outputs are reordered for loss computation. See Appendix for more details.

\subsection{Loss Functions}
After establishing the correspondence between predicted outputs and ground-truth labels via bipartite matching, we optimize the model using a multi-task loss function formulated as:
\begin{equation}
    \mathcal{L} = \lambda_{\text{params}}\mathcal{L}_{\text{params}} + \lambda_{\text{2D}}\mathcal{L}_{\text{2D}} + \lambda_{\text{3D}}\mathcal{L}_{\text{3D}} + \lambda_{\text{box}}\mathcal{L}_{\text{box}},
    \label{eq:total_loss}
\end{equation}
where $\lambda_{\{\cdot\}}$ denotes the weighting coefficients used to balance the respective loss contributions. 

\textbf{Parameter Loss ($\mathcal{L}_{\text{params}}$)} computes the $\ell_2$ distance between the predicted SMAL+ shape and pose parameters and their corresponding ground-truth values if provided.

\textbf{Keypoint Losses ($\mathcal{L}_{\text{2D}}, \mathcal{L}_{\text{3D}}$)} represent the $\ell_1$ distance between the ground-truth 2D and 3D keypoint positions and the ones regressed from predicted SMAL+, respectively.

\textbf{Bounding Box Loss ($\mathcal{L}_{\text{box}}$)} supervises localization accuracy through a combination of coordinate regression, geometric alignment, confidence scoring and the denoising training strategy of DN-DETR~\cite{li2022dn}:
\begin{equation}
    \mathcal{L}_{\text{box}} = \mathcal{L}_{\text{coord}} + \mathcal{L}_{\text{giou}} + \mathcal{L}_{\text{conf}} + 
    \mathcal{L}_{\text{dn}}, 
\end{equation}
where $\mathcal{L}_{\text{coord}}$ is an $\ell_1$ loss over normalized bounding box coordinates, and $\mathcal{L}_{\text{giou}}$ refers to the Generalized IoU (GIoU) loss~\cite{rezatofighi2019generalized}. To refine the objectness score, $\mathcal{L}_{\text{conf}}$ employs a Binary Cross-Entropy (BCE) loss, where the actual IoU between the matched predicted and ground-truth boxes serves as the soft target for the predicted confidence.

%% file: sec/04_dataset.tex
\section{Herd3D Dataset}
\label{sec:dataset}

\begin{figure*}[h] \centering
    % \setlength{\abovecaptionskip}{2pt}
    % \captionsetup{skip=2pt}
    \includegraphics[width=\textwidth]{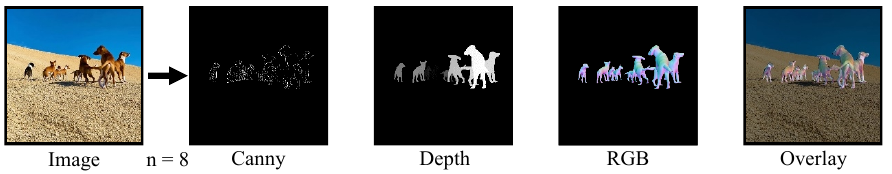}
    \caption{Example from the Herd3D dataset. This figure shows a generated scene with eight dogs, together with the corresponding canny map, depth map, RGB image, and 3D reconstruction overlay. The full Herd3D dataset contains multi-animal scenes with 2 to 8 animals per image.\label{fig:Herd3D}}
    \vspace{-0.3cm}
\end{figure*}

To support multi-animal 3D reconstruction in real-world scenarios, we construct Herd3D, a large-scale dataset specifically designed for multi-animal scenes which contains over 5K images and 118 species (see Fig.~\ref{fig:Herd3D}). We believe that GenZoo~\cite{niewiadomski2025generative} provides a strong and practical starting point for constructing large-scale animal datasets, because it couples a parametric animal model with controllable image synthesis, which enables scalable generation of paired images and geometry while maintaining pose and shape consistency. Building on GenZoo, we adapt the pipeline for multi-animal data generation. To construct group layouts, we sample up to $8$ animals per image and place them on a shared ground plane. For each instance, we set $t_y=0$ and sample translations with $t_x \in [-1.5,1.5]$ using non-adjacent horizontal bins to limit excessive overlap, and $t_z \in [8,50]$ from predefined depth intervals while constraining the group depth span to at most $30$; we add a small $x$ and  $z$ jitter within $\pm1.5$ and apply a fixed ground alignment offset of $0.3$. We further diversify global orientation by sampling pitch in $[-15^\circ,15^\circ]$ and yaw in $[0^\circ,360^\circ]$. To accommodate the increased complexity of multi-animal scenes-including frequent occlusions, higher ambiguity in instance-wise orientation and limited pose diversity, we adapt the GenZoo pipeline with several targeted modifications. We (i) impose scene layout constraints by placing all animals on a shared ground plane, (ii) expand the pose pool by integrating Animal3D poses~\cite{xu2023animal3d} to increase pose diversity, (iii) replace the ControlNet backend with Qwen-Image-ControlNet-Union~\cite{wu2025qwenimagetechnicalreport} to better preserve geometry and occlusion ordering, and (iv) resolve multi-animal orientation ambiguity via a two-stage Qwen3-VL-8B-Instruct~\cite{qwen3technicalreport} prompting scheme, which first predicts left-to-right per-animal facing directions and then composes a single coherent final prompt that integrates the species information, camera settings, and scene attributes. Each synthetic image has a resolution of 1024 × 1024 and includes annotations for SMAL+ parameters, 2D keypoints, 3D keypoints and bounding boxes.

%% file: sec/05_experiment.tex
\section{Experiments}
\noindent\textbf{Datasets.} We curate a comprehensive training corpus of 49.2K images containing both 2D and 3D annotations. Specifically, we aggregate the training splits of Animal Pose~\cite{cao2019cross}, APTv2~\cite{yang2023aptv2}, AwA-Pose~\cite{banik2021novel}, Stanford Extra~\cite{biggs2020left}, Animal3D~\cite{xu2023animal3d}, and our newly introduced Herd3D. For evaluation, following the protocol in AniMer~\cite{lyu2025animer}, we report results on two in-domain datasets (Animal3D and APTv2) alongside an out-of-domain (OOD) dataset Animal Kingdom.

\noindent\textbf{Baselines.} We benchmark our approach against three recent state-of-the-art (SOTA) methods to ensure a comprehensive evaluation across different architectural paradigms. For model-based techniques, we compare with AniMer~\cite{lyu2025animer}, a transformer-based architecture utilizing the SMAL model, and GenZoo~\cite{niewiadomski2025generative}, which builds upon the SMAL+ variant. Additionally, we include 3D Fauna~\cite{li2024learning} as a representative SOTA model-free reconstruction approach.

\noindent\textbf{Evaluation Metrics.} We evaluate 3D accuracy using the Procrustes-Aligned Mean Per Joint Position Error (PA-MPJPE). For 2D accuracy, we report the Percentage of Correct Keypoints (PCK) and keypoint AP/mAP based on Object Keypoint Similarity (OKS)~\cite{lin2014microsoft}. OKS is computed over visible keypoints, with errors normalized by the ground-truth visible-keypoint bounding-box area and $\sigma=0.05$. We report AP$_{50}^{kp}$ and AP$_{75}^{kp}$ at OKS thresholds of 0.50 and 0.75, and mAP$_{50:95}^{kp}$ averaged over thresholds from 0.50 to 0.95 in steps of 0.05.

\noindent\textbf{Implementation Details.} Our network is optimized using AdamW~\cite{loshchilov2017decoupled} with an initial learning rate of $2 \times 10^{-5}$, incorporating a linear warmup over the first 15 epochs. Similar to AniMer~\cite{lyu2025animer}, we employ a two-stage training strategy consisting of 250 epochs for the first stage and 250 epochs for the second. 
We apply prompt dropout for robustness: the mask prompt
is dropped with 50\% probability, the entire keypoint prompt is dropped
with a probability of $p_{\text{drop}}\!=\!0.2$, and otherwise each
keypoint is independently masked out with a rate sampled uniformly
from $[0,\, 0.7]$ per step, encouraging the model to handle partial or
absent prompts at inference time.
Training is distributed across four RTX 4090 GPUs with a gradient accumulation step of 16. To balance the objective function, our empirical loss weighting factors are set to $\lambda_{\text{params}}=1$, $\lambda_{\text{2D}}=5$, $\lambda_{\text{3D}}=5$, and $\lambda_{\text{box}}=1$. 
% The weighting factors in the matching cost (Eq.~\ref{equ:match_loss}) are set as follows: $\lambda_{\text{conf}} = 1$, $\lambda_{\text{bbox}} = 1$, $\lambda_{\text{giou}} = 1$, and $\lambda_{\text{kpts}} = 10$.

\begin{figure*}[t] 
    \centering
    \includegraphics[width=0.8\textwidth]{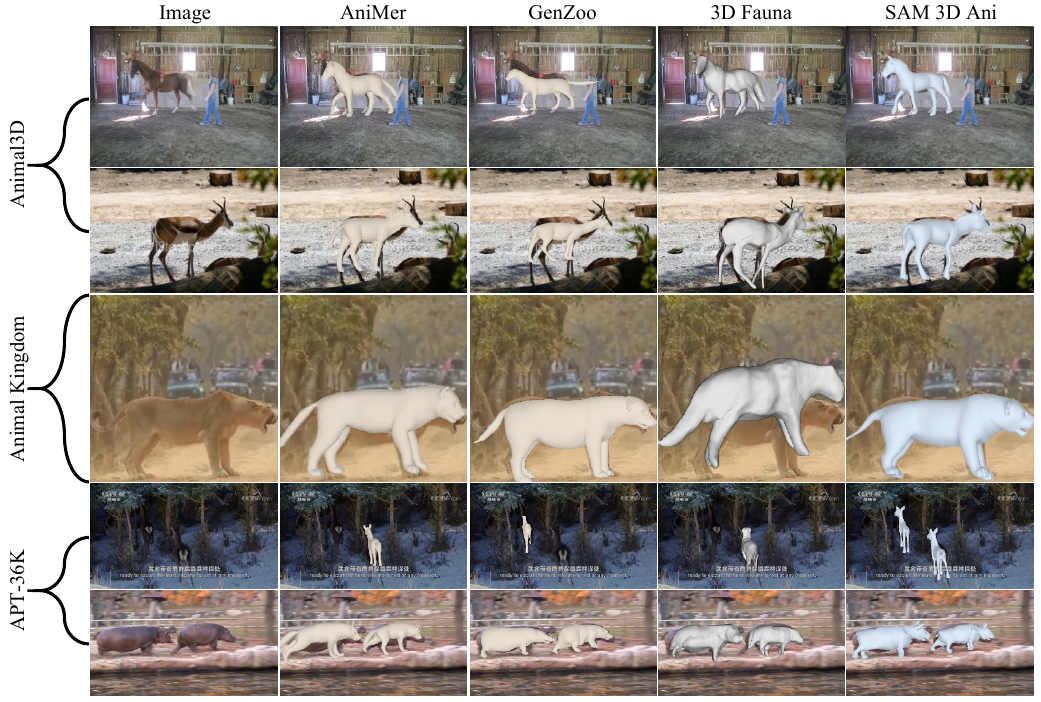}
    \caption{Qualitative comparisons on Animal3D, Animal Kingdom and APT-36K datasets. We compare our results with AniMer~\cite{lyu2025animer}, GenZoo~\cite{niewiadomski2025generative} and 3D Fauna~\cite{li2024learning}. AniMer, GenZoo, and 3D Fauna require pre-cropped single-animal images as input, whereas our method directly processes the full image and retrieves reconstructions whose confidence exceeds a predefined threshold, eliminating the need for cropping. 
    \label{fig:baseline}}
\end{figure*}

\subsection{Comparison}
\noindent\textbf{Comparison without prompts.}
We present the quantitative results in Table~\ref{tab:comparison}. Without any prompt, our method already achieves competitive or superior performance relative to existing approaches.
On Animal3D, the prompt-free variant attains a PA-MPJPE of 80.7\,mm, slightly outperforming AniMer (81.0\,mm), while achieving a higher mAP (49.3 vs.\ 47.2).
On APTv2, keypoint localization improves substantially, with PCK@0.1 reaching 87.9, far surpassing GenZoo (64.1) and AniMer (62.4), though AP remains lower (49.4 vs.\ 55.5 for GenZoo), indicating that the two paradigms exhibit complementary strengths under different metrics.
On the OOD Animal Kingdom benchmark, our prompt-free results lead all metrics, demonstrating stronger generalization to unseen scenes.

\noindent\textbf{Prompt-driven performance gains.}
A key advantage of our framework is its ability to leverage auxiliary prompts at inference time.
When supplied with keypoints from an off-the-shelf ViTPose~\cite{xu2022vitpose} estimator, performance improves consistently: on APTv2, AP rises from 49.4 to 55.5 and mAP from 23.5 to 27.9; on Animal Kingdom, AP increases from 45.0 to 50.5.
This practical variant already matches or surpasses the best baseline on most metrics.
With ground-truth keypoint prompts, the gains become substantially larger: on APTv2, PCK@0.1 reaches 89.0 (vs.\ 62.4 for AniMer) and AP reaches 57.4 (vs.\ 55.5 for GenZoo); on Animal Kingdom, AP improves to 60.1 and mAP to 17.7, roughly doubling AniMer's 10.4.
These results confirm that prompting provides a scalable mechanism for improving reconstruction quality, with performance increasing monotonically as prompt fidelity improves---a unique advantage that existing methods cannot replicate.

\begin{figure*}[t] 
    \centering
    \includegraphics[width=0.55\textheight]{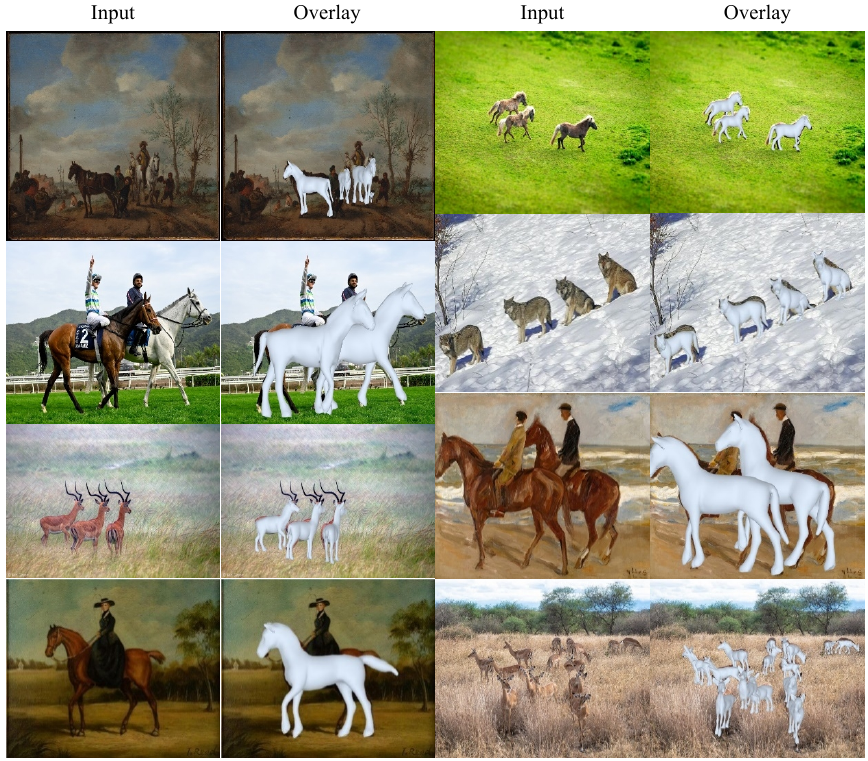}
    \caption{Qualitative evaluation of SAM 3D Animal. For each example, we show: (a) the input image, and (b) the 3D reconstruction overlay. To demonstrate the robustness of SAM 3D Animal, we visualize results across diverse animal species and challenging in-the-wild scenarios, including unusual poses, large viewpoint variations, and crowded scenes with several animals. \label{fig:result}}
\end{figure*}

\noindent\textbf{Qualitative comparison.}
Fig.~\ref{fig:baseline} presents visual comparisons across the three benchmarks.
3D Fauna, as a model-free approach, produces coarse reconstructions that lack geometric detail.
GenZoo and AniMer yield plausible shapes but exhibit less accurate alignment with the input image.
Our method consistently produces reconstructions that are better aligned with the observed pose and viewpoint.
Additional results spanning diverse species and challenging in-the-wild scenarios are shown in Fig.~\ref{fig:result}.

\begin{table}[t]
\centering
\caption{\textbf{Quantitative comparisons on the Animal3D dataset, APTv2 dataset and Animal Kingdom dataset.}}
\label{tab:comparison}
\resizebox{0.9\columnwidth}{!}{
\begin{tabular}{cccccccccc}
\toprule
\multirow{2}{*}{Method} & \multicolumn{3}{c}{Animal3D} & \multicolumn{3}{c}{APTv2} & \multicolumn{3}{c}{Animal Kingdom} \\ 
\cmidrule(lr){2-4} \cmidrule(lr){5-7} \cmidrule(lr){8-10}  
& PA-MPJPE\(\downarrow\) & AP\(\uparrow\) & mAP\(\uparrow\) & PCK@0.1\(\uparrow\) & AP\(\uparrow\) & mAP\(\uparrow\) & PCK@0.1\(\uparrow\) & AP\(\uparrow\) & mAP\(\uparrow\)  \\ \hline
3D Fauna~\cite{li2024learning} & 202.8 & 63.0	& 10.0 & 32.6 & 39.5 & 7.0 & 31.4 & 13.5 & 2.9 \\
GenZoo~\cite{niewiadomski2025generative} & 123.2 & 85.3  & 34.0  & 64.1  & \underline{55.5}  & 19.2  & 42.1  & 23.6  & 4.7  \\
AniMer~\cite{lyu2025animer} & 81.0 & \underline{93.4} & 47.2	& 62.4 & 51.8 & 20.1 & 58.2	& 39.7 & 10.4 \\
\hline
Ours & \underline{80.7} & 92.8 & 49.3 & 87.9 & 49.4 & 23.5 & 59.2 & 45.0 & 12.6 \\
Ours (w/ vitpose) & \underline{80.7} & 92.8 & \underline{50.0} & \underline{88.8} & \underline{55.5} & \underline{27.9} & \underline{63.0} & \underline{50.5} & \underline{14.5} \\
Ours (w/ GT prompt)  & \textbf{75.8} & \textbf{93.8} & \textbf{53.4} & \textbf{89.0} & \textbf{57.4} & \textbf{29.7} & \textbf{66.2} & \textbf{60.1} & \textbf{17.7}       \\ \bottomrule
\end{tabular}
}
\end{table}

\begin{table*}[t]
\centering
\caption{
\textbf{Quantitative comparisons on the APTv2 dataset stratified by the number of animals per image.}}
\label{tab:comparison_animal_number}

\setlength{\tabcolsep}{3.2pt}
\scriptsize
\resizebox{\textwidth}{!}{
\begin{tabular}{c*{15}{c}}
\toprule
\multirow{2}{*}{Method}
& \multicolumn{3}{c}{1 Animal}
& \multicolumn{3}{c}{2 Animals}
& \multicolumn{3}{c}{3 Animals}
& \multicolumn{3}{c}{4 Animals}
& \multicolumn{3}{c}{5+ Animals} \\
\cmidrule(lr){2-4}
\cmidrule(lr){5-7}
\cmidrule(lr){8-10}
\cmidrule(lr){11-13}
\cmidrule(lr){14-16}
& PCK@0.1$\uparrow$ & AP$\uparrow$ & mAP$\uparrow$
& PCK@0.1$\uparrow$ & AP$\uparrow$ & mAP$\uparrow$
& PCK@0.1$\uparrow$ & AP$\uparrow$ & mAP$\uparrow$
& PCK@0.1$\uparrow$ & AP$\uparrow$ & mAP$\uparrow$
& PCK@0.1$\uparrow$ & AP$\uparrow$ & mAP$\uparrow$ \\
\midrule

3D Fauna~\cite{li2024learning}
& 39.8 & 44.6 & 8.4
& 31.5 & 40.4 & 8.0
& 32.5 & 38.1 & 7.8
& 28.7 & 35.6 & 5.7
& 26.8 & 34.8 & 4.5 \\

GenZoo~\cite{niewiadomski2025generative}
& 68.6 & 57.5 & 19.8
& 58.7 & 39.3 & 12.0
& 59.7 & 40.4 & 12.7
& 57.5 & 45.6 & 15.2
& 48.9 & 24.1 & 6.6 \\

AniMer~\cite{lyu2025animer}
& 78.4 & 69.0 & 29.8
& 64.2 & 53.1 & 19.6
& 61.8 & 55.9 & 20.9
& 52.0 & 50.2 & 19.2
& 46.7 & 31.5 & 10.4 \\

\midrule

Ours (w/o GT prompt)
& 83.4 & 81.1 & 42.9
& 64.9 & 50.4 & 21.8
& 62.1 & 48.0 & 21.2
& 42.7 & 31.9 & 12.7
& 28.9 & 13.3 & 4.0 \\

Ours (w/ bbox)
& \textbf{86.8} & \textbf{84.1} & \textbf{46.8}
& \textbf{74.5} & \textbf{63.2} & \textbf{30.0}
& \textbf{78.2} & \textbf{66.5} & \textbf{33.6}
& \textbf{66.9} & \textbf{55.0} & \textbf{26.0}
& \textbf{57.7} & \textbf{37.8} & \textbf{14.9} \\

\bottomrule
\end{tabular}
}
\end{table*}

\noindent\textbf{Performance under Varying Numbers of Animals.}
\label{sec:animal_count}
We further evaluate how reconstruction performance changes with the number of
animals in an image. We stratify the APTv2 test set into images containing
1, 2, 3, 4, and 5+ annotated animals, corresponding to 3,502, 1,384, 557,
166, and 426 images, respectively (6,035 images in total). This grouping
captures increasingly challenging conditions with stronger crowding, occlusion, and smaller animal scale. We report instance-averaged PCK@0.1, AP, and mAP for each group in Table~\ref{tab:comparison_animal_number}. For direct comparison with existing methods, \textit{Ours (w/ bbox)} follows
the same protocol as the baselines: each animal is cropped using its ground-truth bounding box and reconstructed independently. Under this setting, SAM 3D Animal achieves the best performance across all metrics and animal-count groups. For example, it obtains 86.8/84.1/46.8 PCK/AP/mAP on single-animal images and remains strongest for images with 5+ animals, reaching 57.7/37.8/14.9 and outperforming the best baseline by 11.0, 6.3, and 4.5 points, respectively. Without GT prompts, \textit{Ours (w/o GT prompt)} processes the full image and jointly reconstructs all animals without bounding boxes, keypoints, or masks. It remains competitive for single-animal and moderately crowded scenes, but performance decreases more noticeably for 4 and 5+ animals, where jointly detecting and reconstructing all instances becomes substantially more difficult under severe occlusion and scale variation.

\subsection{Ablation}
% w/o our dataset
% w/o prompt (mask, keypoints, mask & keypoints)
% Compared to AniMer with same dataset
\label{sec:Ablation}

\begin{figure*}[t] 
    \centering
    \includegraphics[width=0.6\textheight]{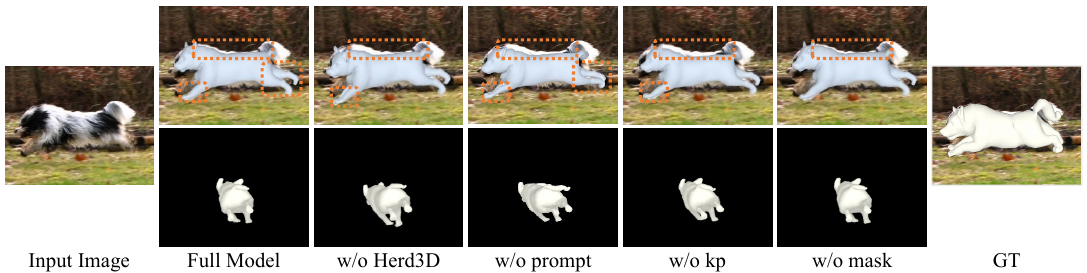}
    \caption{Ablation studies. Keypoint prompting, mask prompting, and training with our Herd3D dataset each lead to improved performance, as discussed in Sec.~\ref{sec:Ablation}. \label{fig:ablation}}
\end{figure*}

\begin{table}[ht]
\centering
\caption{\textbf{Ablation study on the Animal3D dataset, APTv2 dataset and Animal Kingdom dataset.}}
\label{tab:ablation}
\resizebox{0.9\columnwidth}{!}{
\begin{tabular}{cccccccccc}
\toprule
\multirow{2}{*}{Variant} & \multicolumn{3}{c}{Animal3D} & \multicolumn{3}{c}{APTv2} & \multicolumn{3}{c}{Animal Kingdom} \\ 
\cmidrule(lr){2-4} \cmidrule(lr){5-7} \cmidrule(lr){8-10}  
& PA-MPJPE\(\downarrow\) & AP\(\uparrow\) & mAP\(\uparrow\) & PCK@0.1\(\uparrow\) & AP\(\uparrow\) & mAP\(\uparrow\) & PCK@0.1\(\uparrow\) & AP\(\uparrow\) & mAP\(\uparrow\)  \\ \hline
w/o Herd3D & 79.5 & \underline{93.8} & 51.6	& 88.0 & 48.7 & 24.7 & 62.5	& 51.1 & 15.7 \\
w/o prompt & 80.7 &	92.8 & 49.3	& 87.9 & 49.4 & 23.5 & 59.2 & 45.0 & 12.6 \\
w/o kp & 80.8 & 93.1 & 49.8	& 87.9 & 49.7 & 23.5 & 59.3	& 45.4 & 12.8 \\
w/o mask & \textbf{75.8} & \textbf{94.1} & \textbf{53.4} & \underline{88.9} & \underline{57.3} & \underline{29.6} & \underline{66.0} & \textbf{60.1} & \underline{17.5} \\
Full  & \textbf{75.8} & \underline{93.8} & \textbf{53.4} & \textbf{89.0} & \textbf{57.4} & \textbf{29.7} & \textbf{66.2} & \textbf{60.1} & \textbf{17.7} \\ \bottomrule
\end{tabular}
}
\end{table}

\begin{table}[ht]
\centering
\caption{\textbf{Ablation study on the number of prompt keypoints.}}
\label{tab:ablation_kp}
\resizebox{0.9\columnwidth}{!}{
\begin{tabular}{cccccccccc}
\toprule
\multirow{2}{*}{Number} & \multicolumn{3}{c}{Animal3D} & \multicolumn{3}{c}{APTv2} & \multicolumn{3}{c}{Animal Kingdom} \\ 
\cmidrule(lr){2-4} \cmidrule(lr){5-7} \cmidrule(lr){8-10}  
& PA-MPJPE\(\downarrow\) & AP\(\uparrow\) & mAP\(\uparrow\) & PCK@0.1\(\uparrow\) & AP\(\uparrow\) & mAP\(\uparrow\) & PCK@0.1\(\uparrow\) & AP\(\uparrow\) & mAP\(\uparrow\)  \\ \hline
0 & 80.8 & 93.1 & 49.8 & 87.9 & 49.7 & 23.5 & 59.3 & 45.4 & 12.8 \\
5 & 79.5 & \textbf{94.1}	& 51.7 & 88.9 & 54.6 & 26.7	& 63.9 & 56.9 & 16.3 \\
10 & 78.6 & \textbf{94.1} & 51.7 & \textbf{89.1} & 56.8 & 28.7 & 65.3	& 58.6 & 17.1 \\
15  & \textbf{77.4} & \underline{93.8} & \textbf{52.5} & \underline{89.0} & \textbf{57.5} & \textbf{29.6} &	\textbf{66.0} & \textbf{60.2} & \textbf{17.7} \\ \bottomrule
\end{tabular}
}
\end{table}

We ablate three design axes to isolate their respective contributions: training data, prompt modality, and prompt density. Results are reported in Tables~\ref{tab:ablation} and~\ref{tab:ablation_kp}, with qualitative examples in Fig.~\ref{fig:ablation}.

\noindent\textbf{Effect of Herd3D.}
Removing Herd3D from the training set leads to a consistent performance drop across all three benchmarks, with the largest degradation observed on APTv2 (Table~\ref{tab:ablation}).
This is expected: Herd3D is the primary source of multi-animal scenes with per-instance annotations, and its absence disproportionately affects benchmarks that feature crowded or overlapping subjects.
The result validates that our curated dataset fills a genuine gap in existing training resources rather than merely increasing data volume.

\noindent\textbf{Keypoint vs.\ mask prompts.}
Comparing the prompt modality variants in Table~\ref{tab:ablation} reveals a clear asymmetry: keypoint prompts are the dominant contributor, while mask prompts provide only marginal gains.
Specifically, removing the mask prompt (\emph{w/o mask}) leaves performance nearly unchanged from the full model, whereas removing keypoint prompts (\emph{w/o kp}) reduces results to the level of the unprompted baseline.
We attribute this to two factors.
First, keypoints encode the articulated skeletal structure directly, which is precisely what the SMAL model needs to resolve pose ambiguity; masks, by contrast, convey only silhouette-level information that is largely redundant with the image features already extracted by the backbone.
Second, the mask prompts are generated by SAM~\cite{carion2025sam} at inference time, and segmentation errors, particularly on thin limbs or under occlusion, introducing noise that dilutes the prompt signal.
Nonetheless, the full model still edges ahead of the keypoint-only variant on APTv2 and Animal Kingdom, suggesting that mask prompts offer a small but consistent complementary benefit when segmentation quality is adequate.

\noindent\textbf{Number of keypoints.}
Table~\ref{tab:ablation_kp} displays the performance as a function of the number of keypoint prompts, ranging from 0 to 15. For each sample, we randomly select the specified number of keypoints from all available annotations as the prompt input.
Performance improves monotonically, with the steepest gain occurring between 0 and 5 keypoints, even a sparse set of randomly chosen landmarks is sufficient to substantially disambiguate pose.
Beyond 5, improvements continue at a diminishing rate, indicating that the initial keypoints resolve the most salient ambiguities while additional ones refine secondary joints.
This graceful scaling is practically appealing: users can supply as few or as many keypoints as available, and the model extracts value from each additional annotation without requiring a fixed-size input.

\subsection{Impact of Prompt Strategy Under Occlusion}
\label{sec:prompt_occlusion}
\begin{figure}[t]
    \centering
    \includegraphics[width=0.8\linewidth]{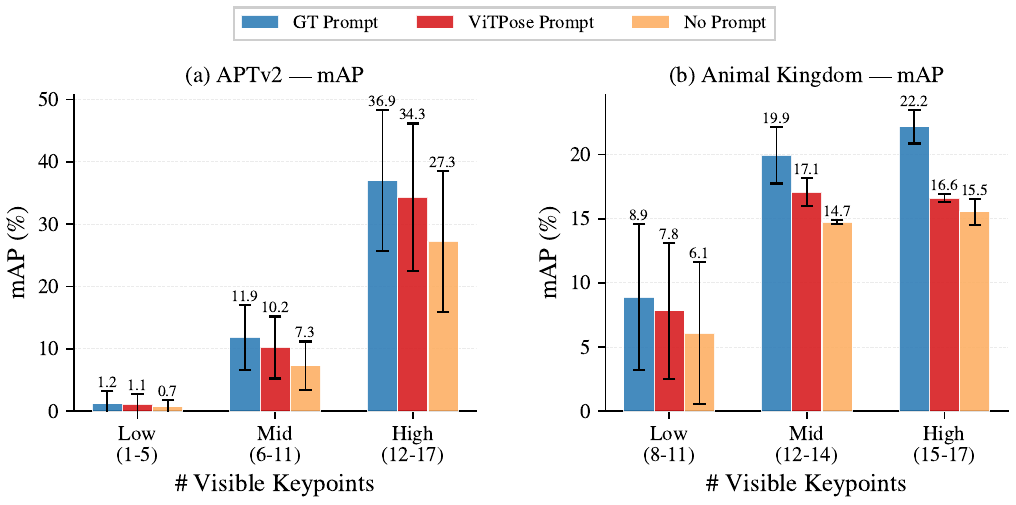}
    \caption{\textbf{Performance under different visibility levels.} We group test samples by the number of visible keypoints into Low, Mid, and High buckets. (a)~mAP on APTv2. (b)~mAP on Animal Kingdom. Error bars denote standard deviation across visibility counts within each group.}
    \label{fig:prompt_occlusion}
\end{figure}

To understand how different prompt strategies behave under varying levels of occlusion, we partition the test set by the number of visible keypoints into three groups (Low, Mid, High) and evaluate three prompt modes:
GT Prompt (ground-truth 2D keypoints), ViTPose Prompt (automatically detected by ViTPose~\cite{xu2022vitpose}), and No Prompt. Results are reported using mAP on both APTv2 and Animal Kingdom, as shown in
Fig.~\ref{fig:prompt_occlusion}. Note that for Animal Kingdom, the lowest visibility group starts at 8--11 keypoints because the dataset's annotations contain a minimum of 8 visible keypoints per instance.

\noindent\textbf{Prompt matters more under heavy occlusion.}
Across both datasets, the relative gain of prompting over the No Prompt baseline is largest in the Low visibility group and diminishes as more keypoints become visible.
On APTv2, ViTPose Prompt improves mAP over No Prompt by 57\% relative in the Low group, 40\% in Mid, and 26\% in High.
On Animal Kingdom, the same trend holds: 28\% relative gain in Low, 16\% in Mid, and 7\% in High.
This confirms that the model relies more heavily on prompt-provided spatial priors when visual evidence is limited, and that prompting offers the greatest practical benefit precisely in the most challenging scenarios.

\noindent\textbf{ViTPose Prompt is a practical alternative to GT.}
On APTv2, ViTPose Prompt consistently performs close to the GT upper bound across all visibility levels, demonstrating that an off-the-shelf keypoint detector can serve as an effective substitute for manual annotations at inference time.

% \noindent\textbf{Coverage outweighs precision under occlusion.}                  
% Most notably, on Animal Kingdom, ViTPose Prompt \emph{surpasses} GT Prompt in the Low visibility group (46.6 vs.\ 38.7). We attribute this to a fundamental difference in prompt coverage: GT Prompt only provides coordinates for the few keypoints that are actually visible, leaving the model with sparse guidance. In contrast, ViTPose can hallucinate plausible locations for occluded keypoints, offering a more complete spatial prior. This suggests that prompt \emph{coverage} (i.e., the number of keypoints with estimated locations) can be more important than prompt \emph{precision} when occlusion is severe.

%% file: sec/06_conclusion.tex
\section{Conclusion}

We presented SAM 3D Animal, a promptable framework for multi-animal 3D reconstruction from a single image. Unlike prior animal reconstruction methods that predominantly focus on animal-centric images, our approach reconstructs multiple animals jointly through a set-prediction formulation and supports flexible keypoint and mask prompts to resolve ambiguity in crowded and occluded scenes. 

\noindent\textbf{Limitation.} While SAM 3D Animal shows strong performance, it remains limited by the SMAL+ shape space and is therefore mainly applicable to quadruped-like animals. Moreover, relative depth ordering between animals is not explicitly constrained, which can cause inaccurate spatial arrangements under severe occlusion. Future work could explore more flexible animal representations, explicit depth-aware scene reasoning, and multi-view refinement by decoupling view-dependent and view-invariant parameters.

%% file: sec/X_suppl.tex
\section{Bipartite Matching Details}
\label{sec:supp_bipartite}
We formulate the assignment between the predicted entities and ground-truth targets as a set prediction problem. Let \(\hat{\mathcal{Y}} = \{\hat{y}_i\}_{i=1}^N\) denote the set of predictions, where each prediction comprises bounding box coordinates, a confidence score, and 2D keypoint locations. Let \(\mathcal{Y} = \{y_j\}_{j=1}^M\) denote the set of ground-truth targets (\(M \leq N\)). Following DETR~\cite{carion2020end}, the assignment is determined by searching for an optimal injection \(\hat{\sigma}: \{1,\dots,M\} \rightarrow \{1,\dots,N\}\) that minimizes the overall bipartite matching cost:
\begin{equation}
    \hat{\sigma} = \underset{\sigma}{\arg\min} \sum_{i=1}^{M} \mathcal{C}_{\text{match}}(\hat{y}_{\sigma(i)}, y_i)
\end{equation}
where \(\sigma(i) \in \{1,\dots,N\}\) is the prediction index assigned to the \(i\)-th ground-truth target, and \(\mathcal{C}_{\text{match}}(\hat{y}_{\sigma(i)}, y_i)\) is the pair-wise assignment cost between prediction \(\sigma(i)\) and ground-truth target \(i\). To ensure that the assigned predictions maintain high localization precision and structural pose alignment, our matching cost is a weighted composition of four distinct terms:
\begin{equation}
    \label{equ:match_loss}
    \mathcal{C}_{\text{match}} = \lambda_{\text{conf}} \mathcal{C}_{\text{conf}} + \lambda_{\text{bbox}} \mathcal{C}_{\text{bbox}} + \lambda_{\text{giou}} \mathcal{C}_{\text{giou}} + \lambda_{\text{kpts}} \mathcal{C}_{\text{kpts}}
\end{equation}
Specifically, the weighting factors in the matching cost (Eq.~\ref{equ:match_loss}) are set as follows: $\lambda_{\text{conf}} = 1$, $\lambda_{\text{bbox}} = 1$, $\lambda_{\text{giou}} = 1$, and $\lambda_{\text{kpts}} = 10$. The terms are defined as follows:
\begin{itemize}
    \item \textbf{Bounding Box Cost (\(\mathcal{C}_{\text{bbox}}\)):} The \(\mathcal{L}_1\) distance between the normalized center coordinates, width, and height of the predicted and ground-truth bounding boxes.
    \item \textbf{GIoU Cost (\(\mathcal{C}_{\text{giou}}\)):} The Generalized Intersection over Union (GIoU) cost~\cite{rezatofighi2019generalized}, which provides a scale-invariant geometric alignment constraint for the bounding boxes.
    \item \textbf{Confidence Cost (\(\mathcal{C}_{\text{conf}}\)):} A focal-style penalty applied to the predicted confidence score \(\hat{c}_i \in [0, 1]\). It is formulated as \(\alpha (1 - \hat{c}_i)^\gamma (-\log \hat{c}_i)\) with focusing parameters \(\alpha=0.25\) and \(\gamma=2.0\) following~\cite{Su_2025_CVPR}, encouraging the model to match ground truth with high-confidence predictions.
    \item \textbf{Keypoint Cost (\(\mathcal{C}_{\text{kpts}}\)):} A pose-aware geometric constraint computed as the mean \(\mathcal{L}_1\) distance between the predicted 2D keypoints and the corresponding ground-truth keypoints, strictly masked to account only for visible joints.
\end{itemize}

The optimal permutation \(\hat{\sigma}\) is computed efficiently using the Hungarian algorithm~\cite{kuhn1955hungarian}. Once the optimal assignment is established, the predicted outputs of multiple animals are reordered to perfectly align with the ground-truth sequence, allowing the subsequent calculation of the standard supervisory loss functions.

\section{More Details about Herd3D}
\begin{figure*}[t] \centering
    % \setlength{\abovecaptionskip}{2pt}
    % \captionsetup{skip=2pt}
    \includegraphics[width=0.9\textwidth]{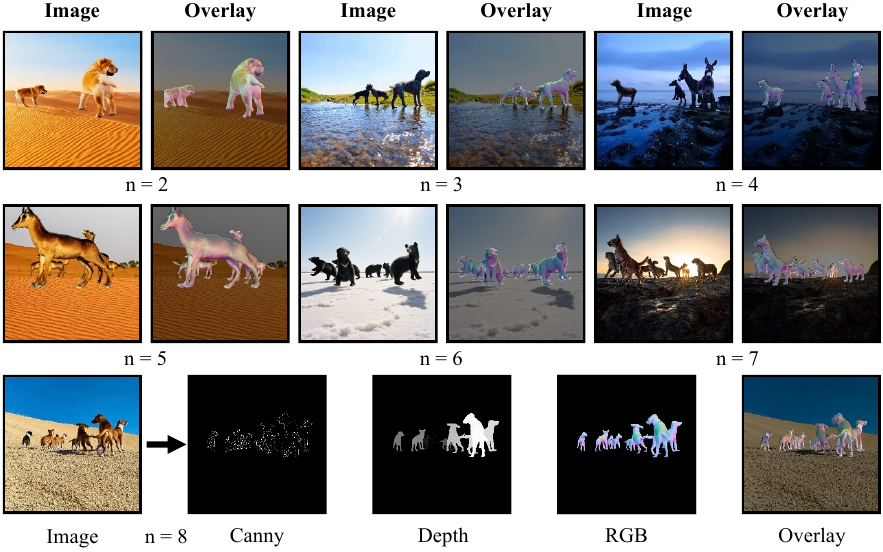}
    \caption{Herd3D multi-animal dataset. The images include dogs, horses, antelopes, bears, and cats, 
with the number of animals per image gradually increasing from 2 to 8 .
\label{fig:Herd3D_full}}
    \vspace{-0.3cm}
\end{figure*}

We provide additional qualitative and statistical details of Herd3D in this section. Fig.~\ref{fig:Herd3D_full} visualizes representative multi-animal examples with varying group sizes, where the number of visible animals gradually increases from two to eight. These examples illustrate the diversity of species, poses, viewpoints, background environment, and occlusion patterns covered by Herd3D, as well as the corresponding structural cues used for annotation and visualization, including edge maps, depth maps, RGB images, and pose overlays. Table~\ref{tab:top10_families} further summarizes the taxonomic composition of the training set by listing the top-10 animal families and representative species within each family. Together, these results show that Herd3D covers both dense multi-animal scenes and broad cross-species diversity, making it suitable for studying scene-centric animal pose understanding.

\section{Failure Cases in Herd3D Construction}
\begin{figure*}[t] \centering
    % \setlength{\abovecaptionskip}{2pt}
    % \captionsetup{skip=2pt}
    \includegraphics[width=0.8\textwidth]{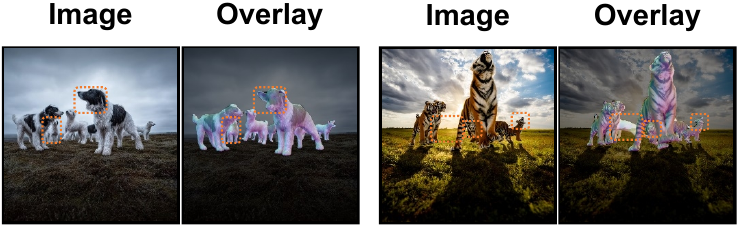}
    \caption{Failure cases of data generation.
\label{fig:Herd3D_failure}}
    \vspace{-0.3cm}
\end{figure*}

During the construction of Herd3D, we observed several failure cases when rendering multi-animal scenes with Qwen-ControlNet. These failures are particularly common when animal meshes have large overlapping regions or severe inter-instance occlusions. In such cases, the rendered image may not faithfully preserve the intended 3D geometry and pose. For example, an animal whose head is oriented away from the camera may be incorrectly rendered with a forward-facing face, leading to inconsistent orientation between the mesh and the generated image. Local semantic errors may also occur, where small body parts are misinterpreted, such as ears being rendered as noses or other facial structures. In addition, when two animals are spatially close, the renderer may blend their body regions, causing the torso or limbs of one animal to be partially rendered onto another. These artifacts indicate that dense multi-animal scenes remain challenging for image-conditioned generative rendering, especially under heavy occlusion and close physical interaction. We provide failure cases in Fig.~\ref{fig:Herd3D_failure}.

\begin{table*}[ht]
\centering
\caption{\textbf{Top-10 animal families in the Herd3D training image set.}
Families are ranked by the number of labelled training images. For each family, we report representative species with their corresponding image counts.}
\label{tab:top10_families}
\scriptsize
\setlength{\tabcolsep}{4pt}
\renewcommand{\arraystretch}{1.08}
\begin{tabular}{c l r p{0.62\textwidth}}
\toprule
Rank & Family & Images & Representative species \\
\midrule
1 & Canidae & 3,238 & Great Dane (105), Curly Coated Retriever (86), Irish Setter (85), Clumber Spaniel (84), Irish Wolfhound (84), Irish Water Spaniel (82), American Cocker Spaniel (73), American Eskimo Dog (71), Cairn Terrier (70), Alaskan Malamute (64) \\
2 & Bovidae & 324 & Cow (106), Cattle (95), American Bison (21), Antelope (21), Goat (21), Bighorn Sheep (20), Buffalo (20), Impala (20) \\
3 & Equidae & 305 & Horse (127), Feral Horse (115), Domestic Horse (21), Donkey (21), Common Zebra (21) \\
4 & Mustelidae & 187 & Honey Badger (73), Eurasian Otter (54), African Clawless Otter (20), American Badger (20), American Mink (20) \\
5 & Rhinocerotidae & 157 & Black Rhinoceros (157) \\
6 & Ursidae & 151 & American Black Bear (90), Brown Bear (21), Da Xiong Mao (20), Giant Panda (20) \\
7 & Felidae & 143 & Bay Lynx (22), European Wildcat (21), Bay Cat (20), Caracal (20), Cat (20), Cheetah (20), Cougar (20) \\
8 & Giraffidae & 63 & Angolan Giraffe (21), Cape Giraffe (21), Giraffe (21) \\
9 & Herpestidae & 42 & Gray Meerkat (42) \\
10 & Viverridae & 41 & Bear Cat (21), African Civet (20) \\
\bottomrule
\end{tabular}
\end{table*}

\section{More Ablation Experiments}
\begin{figure*}[h] 
    \centering
    \includegraphics[width=0.6\textheight]{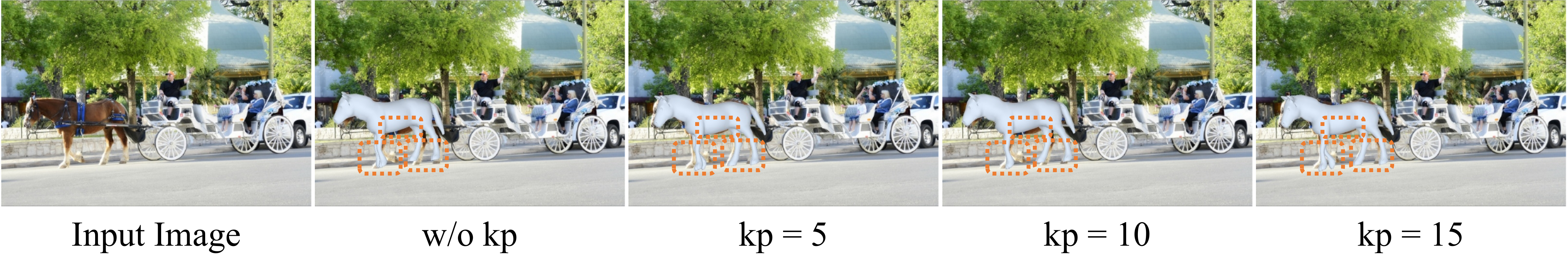}
    \caption{Ablation study on the number of prompt keypoints. \label{fig:ablation_kp}}
\end{figure*}

Fig.~\ref{fig:ablation_kp} further demonstrates the effect on the number of prompt keypoints. With more keypoints used as prompt information, SAM 3D Animal produces results aligned better with input image. 

\section{Inference Efficiency}
\label{sec:inference_efficiency}

We additionally compare the inference efficiency of different methods on APTv2. We group test images by the number of annotated animals in the original image: 1, 2, 3, 4, and 5+ animals, containing 3,502, 1,384, 557, 166, and 426 images, respectively. Table~\ref{tab:inference_time_animal_count} reports the average inference time for each group.

During measurement, we excluded data loading, preprocessing, and metric computation. All measurements were performed on a single NVIDIA RTX 4090 GPU. Note that, for 3D Fauna, GenZoo and AniMer, the input is cropped animal image resized to 256x256 resolution with padding. For SAM 3D Animal, the input is the whole uncropped image resized to 512x512 resolution with padding. Note that, although our method is currently slower than GenZoo and AniMer without any acceleration techniques, it maintains a stable time cost for different numbers of animals, thus having the potential to scale up for inferring a large number of animals efficiently.

\begin{table}[t]
    \centering
    \small
    \caption{
        Forward-pass inference time under different numbers of animals on APTv2.
        The reported time is measured in milliseconds per image and excludes
        data loading, preprocessing, and metric computation.
    }
    \label{tab:inference_time_animal_count}

    \begin{tabular}{cccccc}
        \toprule
        Method
        & 1 Animal
        & 2 Animals
        & 3 Animals
        & 4 Animals
        & 5+ Animals \\
        \midrule
        3D Fauna & 195.9 (195.9) & 198.1 (396.2) & 196.6 (589.9) & 198.1 (792.6) & 195.4 (1518.1) \\
        GenZoo   & 7.8 (7.8)     & 7.8 (15.5)     & 8.0 (24.0)     & 8.0 (31.8)     & 7.8 (60.9) \\
        AniMer   & 7.6 (7.6)     & 7.6 (15.1)     & 7.6 (22.7)     & 7.6 (30.3)     & 7.6 (59.5) \\
        \midrule
        Ours     & 79.4          & 79.2            & 84.7            & 80.3            & 80.5 \\
        \bottomrule
    \end{tabular}
\end{table}

\section{Robustness to Image Quality}
\label{sec:image_quality}
We further study how input image quality affects reconstruction performance on Animal3D. We divide the test set into three approximately balanced groups using an object-centric image-quality score that equally combines percentile-normalized effective animal resolution and crop sharpness. The resulting groups are:
\textit{Low} ($[0, 0.3903]$, 107 images),
\textit{Medium} ($[0.3903, 0.6270]$, 106 images), and
\textit{High} ($[0.6301, 0.9483]$, 107 images).
Table~\ref{tab:image_quality} reports PA-MPJPE, AP, and mAP for each group.

For direct comparison with the baselines, \textit{Ours (w/ bbox)} follows the
same evaluation protocol: each animal is cropped using its ground-truth bounding box and reconstructed independently. In contrast, \textit{Ours (w/o GT prompt)} takes the full image as input and reconstructs the animal without ground-truth bounding boxes, keypoints, or masks.

Image quality primarily affects the 2D reconstruction metrics, while PA-MPJPE
remains relatively stable. In the prompt-free setting, mAP increases from 35.9 on low-quality images to 59.7 on high-quality images. Bounding-box localization substantially improves robustness to degraded images: the low-to-high AP gap is reduced from 12.2 to 6.6 points, while the mAP gap decreases from 23.8 to 15.6 points. These results suggest that accurate localization can partially compensate for limited visual detail, while the bbox-based setting maintains strong performance across all three image-quality levels.

\begin{table}[t]
\centering
\caption{\textbf{Quantitative comparisons on the Animal3D dataset under different image-quality level.}}
\label{tab:image_quality}
\resizebox{0.9\columnwidth}{!}{
\begin{tabular}{cccccccccc}
\toprule
\multirow{2}{*}{Method} & \multicolumn{3}{c}{Low} & \multicolumn{3}{c}{Medium} & \multicolumn{3}{c}{High} \\ 
\cmidrule(lr){2-4} \cmidrule(lr){5-7} \cmidrule(lr){8-10}  
& PA-MPJPE\(\downarrow\) & AP\(\uparrow\) & mAP\(\uparrow\) & PA-MPJPE\(\downarrow\) & AP\(\uparrow\) & mAP\(\uparrow\) & PA-MPJPE\(\downarrow\) & AP\(\uparrow\) & mAP\(\uparrow\)  \\ \hline
3D Fauna~\cite{li2024learning} & 194.7 & 66.3 & 10.5 & 212.8 & 59.2 & 9.4 & 201.0 & 63.4 & 10.1 \\
GenZoo~\cite{niewiadomski2025generative} & 124.8 & 80.4 & 29.8  & 118.6 & 88.7 & 35.5  & 124.3 & 85.0 & 37.3 \\
AniMer~\cite{lyu2025animer} & \textbf{82.6} & 86.3 & 40.8 & 78.1 & 98.5 & 48.8 & 82.3 & 95.5 & 47.2 \\
\hline
% Ours & \underline{80.7} & 92.8 & 49.3 & 87.9 & 49.4 & 23.5 & 59.2 & 45.0 & 12.6 \\
% Ours (w/ vitpose) & \underline{80.7} & 92.8 & \underline{50.0} & \underline{88.8} & \underline{55.5} & \underline{27.9} & \underline{63.0} & \underline{50.5} & \underline{14.5} \\
Ours (w/o GT prompt)
& 84.3 & 85.0 & 35.9
& 79.5 & 96.2 & 52.5
& 78.4 & 97.2 & 59.7 \\

Ours (w/ bbox)
& 83.2 & \textbf{92.5} & \textbf{47.3}
& \textbf{78.0} & \textbf{99.1} & \textbf{57.8}
& \textbf{76.0} & \textbf{99.1} & \textbf{62.9} \\

\bottomrule
\end{tabular}
}
\end{table}

\section{Effect of different types of keypoint prompts.} 
In Table~\ref{tab:animal3d_prompt_locations}, we conduct an additional controlled experiment on Animal3D to investigate whether the
anatomical location of keypoint prompts affects reconstruction quality. To ensure a fair comparison, we select a common subset of
252 images in which all candidate keypoints are visible and provide exactly four keypoint prompts for each image. We compare
Distal keypoints (the four fore- and hind-paw endpoints), Intermediate keypoints (the front elbows and rear knees/hocks), Proximal keypoints (the shoulder and hip keypoints), and Random keypoints (four visible keypoints sampled from all 26 Animal3D keypoints). The Random result is averaged over five seeds, and mask prompts are disabled to isolate the effect of keypoint location.

The results confirm that keypoint selection has a measurable effect. Anatomically selected prompts generally outperform random
prompts. Proximal keypoints achieve the best PA-MPJPE of 75.8, suggesting that shoulder and hip landmarks provide stronger
constraints for global 3D pose recovery. Intermediate keypoints achieve the best mAP of 54.0, indicating that articulated limb
joints are more informative for fine-grained pose estimation. Distal keypoints achieve the best AP of 96.4. Therefore, distal
keypoints are informative, as suggested by the reviewer, but they are not consistently optimal across all metrics. We include this
controlled analysis and the corresponding discussion in the revised manuscript.

\begin{table}[t]
      \centering
      \small
      \caption{
          Effect of keypoint prompt locations on the 252-image common-visible
          subset of Animal3D. \textit{Distal} uses the four fore- and hind-paw
          endpoints; \textit{Intermediate} uses the four intermediate limb
          joints, approximately corresponding to the front elbows and rear
          knees/hocks; \textit{Proximal} uses the four shoulder and hip
          keypoints; and \textit{Random} samples four visible keypoints from
          all 26 Animal3D keypoints.
      }
      \label{tab:animal3d_prompt_locations}
      \begin{tabular}{lccc}
          \toprule
          Prompt type
          & PA-MPJPE $\downarrow$
          & AP $\uparrow$
          & mAP $\uparrow$ \\
          \midrule
          No Prompt    & 77.6 & 95.6          & 51.4          \\
          Random       & 76.4 & 95.4          & 52.3          \\
          Distal       & 76.1 & \textbf{96.4} & 52.5          \\
          Intermediate & 76.0 & 96.0          & \textbf{54.0} \\
          Proximal     & \textbf{75.8} & 95.6 & 53.4          \\
          \bottomrule
      \end{tabular}
\end{table}

\clearpage
\section{Complete Inference Forward Pass}
\label{app:forward_pass}
Algorithm~\ref{alg:sam3d_forward} summarizes the prompt-conditioned forward pass using the indexing of Eqs.~(1)--(4). Here, $\kappa$ denotes image/camera metadata, $\hat{\Theta}$ collects SMAL+ parameters, and $\hat{C}$ denotes camera predictions.

\begin{algorithm}[H]
\caption{SAM 3D Animal inference forward pass ($P=30$, $L=6$)}
\label{alg:sam3d_forward}
\small
\begin{algorithmic}[1]
\Require Image $\boldsymbol{x}$, keypoint prompts $\mathcal{P}$, optional mask $M$, metadata $\kappa$, optional previous estimate $\eta_{\mathrm{prev}}$
\Ensure $P$ scored hypotheses with SMAL+ parameters, cameras, boxes, meshes, and keypoints
\State $\mathcal{F}\gets\operatorname{ImageFeatures}(\boldsymbol{x},M,\kappa)$
\State $\eta_0\gets\operatorname{LearnedInitialEstimate}(P)$
\State $\eta_{\mathrm{prev}}\gets\eta_0$ if no previous estimate is supplied
\State $(\mathcal{Q}_{\mathrm{base}}^{(0)},A_{\mathrm{base}}^{(0)})\gets\operatorname{InitBase}(\eta_0,\eta_{\mathrm{prev}},\kappa)$
\State $\mathcal{P}\gets\operatorname{PadMissingPrompts}(\mathcal{P},c=-2)$
\State $\mathcal{Q}_{\mathrm{prompt}}^{(0)}\gets\operatorname{Concat}_i\bigl(W_p\operatorname{EncodePoint}(p_i)+b_p\bigr)$
\State $\mathcal{Q}^{(0)}\gets\operatorname{Concat}(\mathcal{Q}_{\mathrm{base}}^{(0)},\mathcal{Q}_{\mathrm{prompt}}^{(0)})$ \Comment{Inject prompts once}
\State $A^{(0)}\gets\operatorname{Concat}(A_{\mathrm{base}}^{(0)},\mathcal{Q}_{\mathrm{prompt}}^{(0)})$
\For{$l=0,\ldots,L-1$}
  \State $\mathcal{Q}^{(l+1)}\gets\mathcal{D}_{l+1}(\mathcal{Q}^{(l)},A^{(l)},\mathcal{F})$ \Comment{Eq.~(2)}
  \State $A^{(l+1)}\gets A^{(l)}$
  \State $(\hat{\Theta},\hat{C},\hat{B},\hat{s})^{(l+1)}\gets\operatorname{Predict}(\mathcal{Q}_{\mathrm{params}}^{(l+1)},\mathcal{Q}_{\mathrm{box}}^{(l+1)};\eta_0)$
  \State $(\mathbf{V},\mathbf{X}_{\mathrm{3D}})^{(l+1)}\gets\mathcal{M}(\hat{\Theta}^{(l+1)})$
  \State $(\hat{x}_{\mathrm{2D}},d)^{(l+1)}\gets\operatorname{Project}(\mathbf{X}_{\mathrm{3D}}^{(l+1)},\hat{C}^{(l+1)};\kappa)$
  \If{$l<L-1$} \Comment{Feedback only after the first five blocks}
    \State $m^{(l+1)}\gets\operatorname{ValidProjection}(\hat{x}_{\mathrm{2D}}^{(l+1)},d^{(l+1)})$
    \State $\bar{x}_{\mathrm{3D}}^{(l+1)}\gets\operatorname{CenterPelvis}(\mathbf{X}_{\mathrm{3D}}^{(l+1)})$
    \State $f^{(l+1)}\gets m^{(l+1)}\odot\operatorname{BilinearSample}(\mathcal{F},\hat{x}_{\mathrm{2D}}^{(l+1)})$
    \State $\mathcal{Q}_{\mathrm{2D}}^{(l+1)}\gets\mathcal{Q}_{\mathrm{2D}}^{(l+1)}+\phi_{\mathrm{feat}}(f^{(l+1)})$
    \State $A_{\mathrm{2D}}^{(l+1)}\gets m^{(l+1)}\odot\phi_{\mathrm{pos}}(\hat{x}_{\mathrm{2D}}^{(l+1)})$ \Comment{Eq.~(3)}
    \State $A_{\mathrm{3D}}^{(l+1)}\gets\psi_{\mathrm{pos}}(\bar{x}_{\mathrm{3D}}^{(l+1)})$ \Comment{Eq.~(4)}
  \EndIf
\EndFor
\State \Return $\{\hat{\Theta}_j^{(L)},\hat{C}_j^{(L)},\hat{B}_j^{(L)},\hat{s}_j^{(L)},\mathbf{V}_j^{(L)},\mathbf{X}_{\mathrm{3D},j}^{(L)},\hat{x}_{\mathrm{2D},j}^{(L)}\}_{j=1}^{P}$
\end{algorithmic}
\end{algorithm}

\noindent\textbf{Initialization and prompting.}
$\operatorname{ImageFeatures}$ extracts ViT features with the configured mask and ray conditioning. $\operatorname{InitBase}$ uses learned parameters to construct the four base query groups and the previous-estimate conditioning tokens included in the parameter-related inputs. Their initial positional augmentations contain the previous-estimate embeddings, with zero augmentation for the box and keypoint groups. $\operatorname{EncodePoint}$ is the piecewise encoding defined in Section~3.2: valid keypoint IDs use position-plus-label embeddings, while labels $-1$ and $-2$ use the negative and invalid embeddings. Missing prompts are represented by invalid entries.

\noindent\textbf{Predictions and feedback.}
$\operatorname{Predict}$ includes decoder-output normalization and the SMAL+, camera, box, and confidence heads. $\operatorname{Project}$ produces crop-image coordinates and depths; $m$ selects in-frame projections with valid positive depth. The mask is applied before feature projection and after positional encoding, as in the implementation. Feedback uses the keypoint subsets associated with the 2D and 3D token groups, and preserves 3D token contents and all unlisted components. All hypotheses are returned with confidence scores; a configured confidence threshold can select output instances. During training, denoising queries, auxiliary supervision, and bipartite matching supplement this inference path as described in Section~3.3 and Appendix~A.

%% file: checklist.tex
\section*{NeurIPS Paper Checklist}

%%% BEGIN INSTRUCTIONS %%%
The checklist is designed to encourage best practices for responsible machine learning research, addressing issues of reproducibility, transparency, research ethics, and societal impact. Do not remove the checklist: {\bf The papers not including the checklist will be desk rejected.} The checklist should follow the references and follow the (optional) supplemental material.  The checklist does NOT count towards the page
limit. 

Please read the checklist guidelines carefully for information on how to answer these questions. For each question in the checklist:
\begin{itemize}
    \item You should answer \answerYes{}, \answerNo{}, or \answerNA{}.
    \item \answerNA{} means either that the question is Not Applicable for that particular paper or the relevant information is Not Available.
    \item Please provide a short (1--2 sentence) justification right after your answer (even for \answerNA). 
   % \item {\bf The papers not including the checklist will be desk rejected.}
\end{itemize}

{\bf The checklist answers are an integral part of your paper submission.} They are visible to the reviewers, area chairs, senior area chairs, and ethics reviewers. You will also be asked to include it (after eventual revisions) with the final version of your paper, and its final version will be published with the paper.

The reviewers of your paper will be asked to use the checklist as one of the factors in their evaluation. While \answerYes{} is generally preferable to \answerNo{}, it is perfectly acceptable to answer \answerNo{} provided a proper justification is given (e.g., error bars are not reported because it would be too computationally expensive'' or ``we were unable to find the license for the dataset we used''). In general, answering \answerNo{} or \answerNA{} is not grounds for rejection. While the questions are phrased in a binary way, we acknowledge that the true answer is often more nuanced, so please just use your best judgment and write a justification to elaborate. All supporting evidence can appear either in the main paper or the supplemental material, provided in appendix. If you answer \answerYes{} to a question, in the justification please point to the section(s) where related material for the question can be found.

IMPORTANT, please:
\begin{itemize}
    \item {\bf Delete this instruction block, but keep the section heading ``NeurIPS Paper Checklist"},
    \item  {\bf Keep the checklist subsection headings, questions/answers and guidelines below.}
    \item {\bf Do not modify the questions and only use the provided macros for your answers}.
\end{itemize}

%%% END INSTRUCTIONS %%%

\begin{enumerate}

\item {\bf Claims}
    \item[] Question: Do the main claims made in the abstract and introduction accurately reflect the paper's contributions and scope?
    \item[] Answer: \answerYes{} % Replace by \answerYes{}, \answerNo{}, or \answerNA{}.
    \item[] Justification: The abstract and introduction clearly state the main contributions: SAM 3D Animal, a promptable multi-animal 3D reconstruction framework; Herd3D, a multi-animal 3D dataset with per-instance SMAL+ annotations; and empirical comparisons on Animal3D, APTv2, and Animal Kingdom. These claims are supported by the quantitative comparisons, ablation studies, and prompt analysis reported in the experiments.

\item {\bf Limitations}
    \item[] Question: Does the paper discuss the limitations of the work performed by the authors?
    \item[] Answer: \answerYes{} % Replace by \answerYes{}, \answerNo{}, or \answerNA{}.
    \item[] Justification: The conclusion discusses two main limitations: the method is constrained by the SMAL+ shape space and is therefore mainly applicable to quadruped-like animals, and relative depth ordering between animals is not explicitly constrained under severe occlusion.


\item {\bf Theory assumptions and proofs}
    \item[] Question: For each theoretical result, does the paper provide the full set of assumptions and a complete (and correct) proof?
    \item[] Answer: \answerNA{} % Replace by \answerYes{}, \answerNo{}, or \answerNA{}.
    \item[] Justification: The paper does not present theoretical results, theorems, or formal proofs. The mathematical formulations are used to describe the model architecture, bipartite matching, and loss functions.


    \item {\bf Experimental result reproducibility}
    \item[] Question: Does the paper fully disclose all the information needed to reproduce the main experimental results of the paper to the extent that it affects the main claims and/or conclusions of the paper (regardless of whether the code and data are provided or not)?
    \item[] Answer: \answerYes{} % Replace by \answerYes{}, \answerNo{}, or \answerNA{}.
    \item[] Justification: The paper describes the model architecture, Herd3D generation process, training datasets, evaluation datasets, metrics, optimizer, learning rate, training schedule, prompt dropout strategy, loss weights, and compute setup. The supplementary material further provides details of the bipartite matching formulation and experiment results.


\item {\bf Open access to data and code}
    \item[] Question: Does the paper provide open access to the data and code, with sufficient instructions to faithfully reproduce the main experimental results, as described in supplemental material?
    \item[] Answer: \answerNo{} % Replace by \answerYes{}, \answerNo{}, or \answerNA{}.
    \item[] Justification: The code will be made publicly available upon paper acceptance.

\item {\bf Experimental setting/details}
    \item[] Question: Does the paper specify all the training and test details (e.g., data splits, hyperparameters, how they were chosen, type of optimizer) necessary to understand the results?
    \item[] Answer: \answerYes{} % Replace by \answerYes{}, \answerNo{}, or \answerNA{}.
    \item[] Justification: The paper specifies the training corpus, evaluation datasets, baselines, metrics, optimizer, initial learning rate, warmup schedule, training epochs, prompt dropout strategy, gradient accumulation, GPU setup, and loss weights. These details are sufficient to understand the reported experimental results.


\item {\bf Experiment statistical significance}
    \item[] Question: Does the paper report error bars suitably and correctly defined or other appropriate information about the statistical significance of the experiments?
    \item[] Answer: \answerNo{} % Replace by \answerYes{}, \answerNo{}, or \answerNA{}.
    \item[] Justification: The paper reports error bars for the visibility-level analysis in Fig. 8, where the error bars denote standard deviation across visibility counts within each group. However, the main quantitative comparison and ablation tables do not report repeated-run error bars or statistical significance tests because running inference and evaluation is computationally expensive due to optimization.


\item {\bf Experiments compute resources}
    \item[] Question: For each experiment, does the paper provide sufficient information on the computer resources (type of compute workers, memory, time of execution) needed to reproduce the experiments?
    \item[] Answer: \answerYes{} % Replace by \answerYes{}, \answerNo{}, or \answerNA{}.
    \item[] Justification: The paper reports that training is distributed across four NVIDIA RTX 4090 GPUs with gradient accumulation in experiment section.

    
\item {\bf Code of ethics}
    \item[] Question: Does the research conducted in the paper conform, in every respect, with the NeurIPS Code of Ethics \url{https://neurips.cc/public/EthicsGuidelines}?
    \item[] Answer: \answerYes{} % Replace by \answerYes{}, \answerNo{}, or \answerNA{}.
    \item[] Justification: The work focuses on 3D animal reconstruction from public and synthetic animal imagery. It does not involve human subjects, personally identifiable information, or safety-critical deployment.


\item {\bf Broader impacts}
    \item[] Question: Does the paper discuss both potential positive societal impacts and negative societal impacts of the work performed?
    \item[] Answer: \answerYes{} % Replace by \answerYes{}, \answerNo{}, or \answerNA{}.
    \item[] Justification: Potential positive impacts include wildlife monitoring, animal behavior analysis, and digital content creation as mentioned in introduction, while potential negative impacts may include misuse for intrusive animal surveillance or inaccurate downstream analysis.

    
\item {\bf Safeguards}
    \item[] Question: Does the paper describe safeguards that have been put in place for responsible release of data or models that have a high risk for misuse (e.g., pre-trained language models, image generators, or scraped datasets)?
    \item[] Answer:\answerNA{} % Replace by \answerYes{}, \answerNo{}, or \answerNA{}.
    \item[] Justification: The proposed model and dataset are not high-risk dual-use assets such as language models, biometric identification systems, or unsafe scraped datasets. The new dataset is generated through a controlled synthetic pipeline.


\item {\bf Licenses for existing assets}
    \item[] Question: Are the creators or original owners of assets (e.g., code, data, models), used in the paper, properly credited and are the license and terms of use explicitly mentioned and properly respected?
    \item[] Answer: \answerYes{} % Replace by \answerYes{}, \answerNo{}, or \answerNA{}.
    \item[] Justification: The paper cites the existing datasets, models, and methods used, including SMAL+, GenZoo, Animal3D, APTv2, Animal Kingdom, ViTPose, and Qwen-based components.

\item {\bf New assets}
    \item[] Question: Are new assets introduced in the paper well documented and is the documentation provided alongside the assets?
    \item[] Answer: \answerYes{} % Replace by \answerYes{}, \answerNo{}, or \answerNA{}.
    \item[] Justification: The paper introduces Herd3D and documents its generation process, scene layout constraints, animal placement strategy, pose sources, ControlNet backend, orientation prompting scheme, image resolution, and available annotations.


\item {\bf Crowdsourcing and research with human subjects}
    \item[] Question: For crowdsourcing experiments and research with human subjects, does the paper include the full text of instructions given to participants and screenshots, if applicable, as well as details about compensation (if any)? 
    \item[] Answer: \answerNA{} % Replace by \answerYes{}, \answerNo{}, or \answerNA{}.
    \item[] Justification: The paper does not involve crowdsourcing experiments or research with human subjects.


\item {\bf Institutional review board (IRB) approvals or equivalent for research with human subjects}
    \item[] Question: Does the paper describe potential risks incurred by study participants, whether such risks were disclosed to the subjects, and whether Institutional Review Board (IRB) approvals (or an equivalent approval/review based on the requirements of your country or institution) were obtained?
    \item[] Answer: \answerNA{} % Replace by \answerYes{}, \answerNo{}, or \answerNA{}.
    \item[] Justification: The paper does not involve human subjects, user studies, or crowdsourced human participant experiments, so IRB approval is not applicable.


\item {\bf Declaration of LLM usage}
    \item[] Question: Does the paper describe the usage of LLMs if it is an important, original, or non-standard component of the core methods in this research? Note that if the LLM is used only for writing, editing, or formatting purposes and does \emph{not} impact the core methodology, scientific rigor, or originality of the research, declaration is not required.
    %this research? 
    \item[] Answer: \answerYes{} % Replace by \answerYes{}, \answerNo{}, or \answerNA{}.
    \item[] Justification: The paper describes the use of Qwen3-VL-8B-Instruct in the Herd3D generation pipeline to resolve multi-animal orientation ambiguity, and also describes the use of Qwen-Image-ControlNet-Union as the image generation backend. These uses are part of the dataset generation method rather than only writing or editing assistance.


\end{enumerate}